# Towards interpretable models for language proficiency assessment: Predicting the CEFR level of Estonian learner texts

Kais Allkivi
Tallinn University

**Abstract.** Using NLP to analyze authentic learner language helps to build automated assessment and feedback tools. It also offers new and extensive insights into the development of second language production. However, there is a lack of research explicitly combining these aspects. This study aimed to classify Estonian proficiency examination writings (levels A2–C1), assuming that careful feature selection can lead to more explainable and generalizable machine learning models for language testing. Various linguistic properties of the training data were analyzed to identify relevant proficiency predictors associated with increasing complexity and correctness, rather than the writing task. Such lexical, morphological, surface, and error features were used to train classification models, which were compared to models that also allowed for other features. The pre-selected features yielded a similar test accuracy but reduced variation in the classification of different text types. The best classifiers achieved an accuracy of around 0.9. Additional evaluation on an earlier exam sample revealed that the writings have become more complex over a 7–10-year period, while accuracy still reached 0.8 with some feature sets. The results have been implemented in the writing evaluation module of an Estonian open-source language learning environment.

## 1. Introduction

An important application of authentic language material in AI-assisted language learning is the development of writing assessment systems. In second language (L2) assessment, authenticity usually refers to test tasks that reflect real-world social situations (Gilmore, 2019). This relates to the focus on communicative language competence promoted by the Common European Framework of Reference for Languages (CEFR). Although the test setting requires learners to display maximal L2 knowledge under a time constraint, their language production can still be considered authentic as they engage in communicative tasks (Ellis & Barkhuizen, 2005; Gilmore, 2019). Writing assessment tools trained on test or examination data can therefore also be useful for evaluating the proficiency of texts written in learning or daily situations.

In addition to high-stakes grading, writing assessment applications can facilitate course placement testing, provide instant feedback to L2 learners, and support teachers in checking written assignments, allowing them to monitor student progress. Research on automatic CEFR level prediction has contributed to the development of language learning software, like Write & Improve for English (Yannakoudakis et al., 2018), Lärka for Swedish (Pilán, 2018), and EVALD for Czech (Rysová et al., 2019). With the broader goal of supporting similar (self-)assessment opportunities in learning and teaching the Estonian language, this study aimed to find optimal solutions for classifying Estonian learner writings on the CEFR scale. The task was limited to the levels A2–C1, which are tested in national exams and can be required for work.

Automated language testing has gained relevance in Estonia due to the planned digitization of state exams for upper secondary school graduates (12th grade), final exams for basic school graduates (9th grade), and language proficiency exams. CEFR-based classification, intended for general writing practice and assessment purposes, is not directly suitable for scoring specific exam writing tasks. However, it helps uncover relevant associations between different levels and characteristics of L2 usage, thus serving as a basis for more specialized assessment solutions. A further aim was to demonstrate how combining machine learning with corpus analysis and careful selection of reliable linguistic features can help ensure the interpretability and robustness of automated proficiency level profiling.

The following research question was stated: How effectively can Estonian A2–C1-level writings be classified using different types of linguistic features that prove to be relevant predictors of proficiency level? The language material included various types of creative writings produced in Estonian language proficiency exams. Four categories of features were extracted from the texts: lexical features – variables of lexical complexity; morphological features – frequencies of parts-of-speech and grammatical categories of nominals and verbs; surface features – text complexity measures related to word, sentence, and text length; error features – frequencies of spelling and grammatical errors as detected by correction tools. Reliable predictors were defined based on several criteria: significant difference between all or some adjacent levels (A2–B1, B1–B2, B2–C1), monotonic increase or decrease in feature values, correlation to proficiency level, and no substantial variation due to text type.

For each feature category, classifiers were trained under two conditions, allowing only pre-selected or all available features to be contained in the models. In addition to overall accuracy, per-level recall and precision, the similarity in classifying different text types was considered in model evaluation. Two test sets – one drawn from a different sample of exam writings – were used to validate both within- and cross-dataset generalizability.

The best-performing model parameters have been implemented in the Writing Evaluator tool of the Estonian language learning and analysis environment ELLE (Allkivi et al., 2024). The online platform integrates the Estonian Interlanguage Corpus (EIC) and text analysis software. Proficiency assessment is accompanied by spelling and grammatical error correction, feedback on text complexity, and editing suggestions (e.g., pointing out long sentences and word repetitions).

**2. Related work**

While the CEFR (2001; 2020) provides a unified description of language competence development, the features of successful language use at different levels are largely language dependent. Identifying these "criterial" features from human-graded learner corpus material (see O'Keeffe & Mark, 2017; Wisniewski, 2017) closely relates to the problem of CEFR-based writing assessment when solved by supervised machine learning. The dominant approach has been the use of multiclass text classification (or ranking) with various quantitative measures, including lexical, grammatical, discourse, error-based, and simple textual complexity features (Hancke, 2013; Tack et al., 2017; Pilán, 2018; Vajjala & Rama, 2018; Yannakoudakis et al., 2018; Ballier et al., 2019; Rysová et al., 2019; Szügyi et al., 2019; Kerz et al., 2021; Gaillat et al., 2022; Ribeiro-Flucht et al., 2024; Glišić et al., 2025). An alternative is to apply neural classifiers based on BERT embeddings (Schmalz & Brutti, 2022; Lagutina et al., 2025), but they have proven less effective than feature-based models on smaller datasets consisting of 1,000 or fewer texts (Vajjala & Rama, 2018; Caines & Buttery, 2020; Glišić et al., 2025).

Recent experiments have made use of pretrained large language models (LLMs), either fine-tuning BERT-like models (Wilkens et al., 2023; Muñoz Sánchez et al., 2024) or prompting generative transformers by providing a scoring rubric and optional rating examples (Naismith et al., 2023; Yancey et al., 2023; Bannó et al., 2024; Yamashita, 2024; Benedetto et al., 2025). This opens new possibilities for low-resource settings and grading pragmatic competence, which has been more challenging for traditional methods, while raising questions about the transparency and trustworthiness of automated assessment. Despite the efforts to explain the functioning and responses of deep learning models (e.g., Bannó et al., 2024; Muñoz Sánchez et al., 2024), the decisions made by statistical machine learning models remain easier to interpret and enable evidence-based feedback on learner performance. The feature-based and LLM-based approaches are complementary and can be used to develop hybrid systems for writing assessment (e.g., Atkinson & Palma, 2025; Hou et al., 2025).

The present study attempted to further enhance the interpretability of feature-based supervised learning by combining CEFR level prediction with learner corpus analysis. Corpus-informed pre-selection of relevant linguistic features was assumed to improve the performance and transparency of prediction models, enabling learners and teachers to understand what is graded and how it relates to writing proficiency development. Previous work presents little information about significant differences between proficiency levels, primarily highlighting some of the best predictive variables according to feature selection ranking or their importance in the model. A somewhat more comprehensive insight is offered by Gaillat et al. (2022) and Ribeiro-Flucht et al. (2024), who describe level-to-level changes in the most informative features of their model, as well as Wilkens et al. (2023), who reveal a variety of features correlated with proficiency level and the distribution of top correlated feature values. Nonetheless, it remains mostly unclear which features exactly differentiate which levels and how they vary across text types.

The same applies to the Estonian context. An earlier experiment by Vajjala and Lõo (2014) to classify Estonian A2–C1-level texts based on lexical and morphological features yielded promising results with a prediction accuracy of 0.79. However, some of their highest-ranked distinguishing features (2nd-person verb forms, subordinating conjunctions, and interjections) have turned out to be task-dependent (Allkivi-Metsoja, 2021). To achieve reusability across different text types, Yannakoudakis et al. (2018) suggest using predictors that associate with general writing competence rather than the text topic or genre. The criteria for detecting such reliable features can be drawn from previous lexical and grammatical analysis of Estonian L2 exam writings (Allkivi-Metsoja, 2021; 2022).

Other related work includes the development of a tool that rates the proficiency level of Estonian words on the scale from pre-A1 to C1, but does not evaluate the text as a whole (Üksik et al., 2021), and the experiments of scoring Estonian L1 essays from 9th- and 12th-grade pilot e-exams (Karjus et al., 2026). The latter show that the zero-shot prompting of LLMs and the statistical NLP-based approach complement each other and can both achieve performance comparable to that of human raters.

## 3. Materials and methods

### 3.1 Proficiency examination data

The main dataset contains 720 Estonian language proficiency examination writings from the years 2017–2020. Each proficiency level is represented by 180 texts graded with at least 60% of the maximum score and thus meeting the level requirements. A2–B2-level material has been produced in six exams, four of them held in 2018 and two in 2020. A random sample of 30 texts was selected per exam. Due to fewer participants, the C1-level material was randomly sampled from 12 exams, including four exams held in 2017, 2018, and 2020. The number of writings could not be balanced and follows their actual distribution between exams. The originally handwritten texts, along with anonymous metadata, were provided by the Estonian Education and Youth Board after digitization and pseudonymization, during which personal names and contact information were manually replaced. The material is publicly available in the L2 proficiency exam subcorpus of the EIC[1].

Although the writing part of the exams consists of two tasks, the creative writing tasks were chosen for the analysis. The tasks involved various text types. At level A2, these were messages to an acquaintance (e.g., asking for help), categorized here as personal letters, or narrative/descriptive texts (e.g., about the last trip). B1 texts were also either personal letters (e.g., sharing impressions from a concert visit) or narrative texts (e.g., about keeping pets). From B2, three types of texts were included: argumentative writings (e.g., about distance education), personal letters (e.g., asking advice on resolving a work problem), and semi-formal letters (e.g., a complaint to the local government). The letters were selected when the second task was a statistics-based report that relied heavily on source material. C1 texts were argumentative writings to the public or colleagues (e.g., discussing the importance of waste sorting or digital competences).

The age of the examinees ranged from 12 to 70 years. Their mean age was 32.05 years (*SD* = 12.7). The largest proportion (40.6%) of the L2 learners had higher education. 20.6% had primary or basic, 18.75% secondary specialized or vocational, and 16.7% secondary education. 71.7% were female and 27.6% male. No information is collected about the examinees' language background; however, some assumptions can be made based on their citizenship. The sample represents language users with 34 different citizenships. 54.9% were Estonian, 18.3% Russian, 4.7% Ukrainian, 2.2% Latvian, and 2.1% Finnish citizens. 9.6% had undefined citizenship. Since 29% of the Estonian population speaks Russian as L1 (Population census, 2022), the text data is mainly representative of Russian-speaking learners of Estonian. The examinees came from 37 Estonian municipalities. 44.6% resided in the capital, Tallinn. 18.2% lived in Narva and 8.5% in Kohtla-Järve – northeastern cities with a Russian-speaking majority.

---

[1] Corpus query is available at https://elle.tlu.ee/tools/wordlist.

This sample was divided into a training and test set consisting of 600 and 120 texts, respectively (see Table 1 for their text type distribution). The training set was used for comparative linguistic analysis of proficiency levels, cross-validation of classification parameters, and building the final machine learning models. These were tested on the holdout test set stratified for level and text types (henceforth referred to as test set 1).

**Table 1.** Distribution of the 600 training and 120 test texts in the main dataset

| Proficiency level | Training set | | Test set | |
|---|---|---|---|---|
| | Text type | Freq. | Text type | Freq. |
| A2 | personal letter | 105 | personal letter | 15 |
| | narrative text | 45 | narrative text | 15 |
| B1 | personal letter | 75 | personal letter | 15 |
| | narrative text | 75 | narrative text | 15 |
| B2 | personal letter | 50 | personal letter | 10 |
| | semi-formal letter | 50 | semi-formal letter | 10 |
| | argumentative text | 50 | argumentative text | 10 |
| C1 | argumentative text | 150 | argumentative text | 30 |

Additional evaluation of model generalizability employed a separate sample of exam writings (henceforth test set 2). It comprises 398 texts from 2010, representing material from one exam per proficiency level and therefore fewer text types. The sample retrieved from the EIC contains 115 A2- and 109 B1-level narrative writings, 63 B2-level semi-formal letters and argumentative writings (from the same authors), and 49 C1-level argumentative writings. Text length requirements were similar to the newer exam writing tasks: at least 30 words for level A2, approximately 100 words for B1, 140 words for B2 letters, 180 for B2 argumentative texts, and 250 words (now 220–260 words) for C1. No metadata was available about the text authors.

It must be noted that both the CEFR-based grading principles and exam tasks have been consistently developed since the six-level language proficiency scale was adopted in Estonia in 2008. As disclosed in personal correspondence with experts involved in organizing Estonian language proficiency testing, special effort has been dedicated to improving the validity of C1-level writing tasks. Revised task and grading instructions were introduced in 2017, setting a greater emphasis on the skills of writing a well-structured argumentative text. However, as no other open-access L2 exam data were available, test set 2 was considered valuable reference material despite the differences from the main dataset.

## 3.2 Analysis of linguistic features

The texts were tokenized, sentence-segmented, lemmatized, and morphologically tagged with the Stanza Python package (Qi et al., 2020) to extract lexical, morphological, and surface features. Error features were derived from the output of a context-sensitive statistical spell-checker (Allkivi-Metsoja & Kippar, 2023) and a machine-translation-based grammar correction tool (Luhtaru et al., 2024). These features represent two dimensions of L2 performance: accuracy, or correctness, and complexity, which refers to elaborateness and diversity of language production (Housen & Kuiken, 2009). In terms of communicative language competence, as defined in the CEFR (2020), this work relates to the following aspects of linguistic competence: general linguistic range, vocabulary range, grammatical accuracy, vocabulary control, and orthographic control.

**Morphological features** ($n = 111$), forming the largest feature group, characterize 1) the distribution of parts-of-speech (PoS) and their subcategories, such as coordinating/subordinating conjunctions or pre- and postpositions, and 2) the usage of grammatical categories in verbs and nominal words (nouns, adjectives, pronouns, and numerals)[2]. Verb features express the tense, person, number, mood, voice, and polarity of finite verbs and the form of non-finite verbs. Nominal features were calculated in total and individually for nouns, adjectives, and pronouns. The grammatical categories include case, number, and degree of comparison in adjectives. Alongside the 14 cases traditionally distinguished in Estonian, Stanza detects the short (additive) form of the illative case. Singular and plural forms were considered duplicate features for nouns and pronouns, but counted separately for verbs and adjectives, which respectively involve non-finite forms and genitive attributes that do not show number.

**Lexical features** ($n = 20$) represent different aspects of lexical complexity, namely lexical diversity, sophistication, and density (see Read, 2000; Lu, 2012). The latter denotes a single measure, i.e., the proportion of content words, whereas 13 diversity and six sophistication features were analyzed.

Lexical diversity can most simply be measured as the number of unique words in text or its ratio to the total number of words, known as the type-token ratio (TTR). In addition, the following standardized indices were used: root type-token ratio (RTTR; Guiraud, 1960) and Uber index (Dugast, 1978) previously applied to compare Estonian A2–C1-level texts (Alp et al., 2013; Vajjala & Lõo, 2014) or L2 and L1 texts (Kerge & Pajupuu, 2009), as well as the Maas index (Maas, 1972) and the Measure of Textual Lexical Diversity (MTLD; McCarthy & Jarvis, 2010), because they capture unique lexical information and show little dependence on text length. PoS-specific TTRs and corrected verb variation (CVV) were also calculated, following Lu (2012) and

---

[2] The guide of the Estonian universal dependencies tagset can be found at https://universaldependencies.org/et/.

Vajjala and Lõo (2014). Due to Estonian rich morphology, lemmas instead of word forms were considered as different words (types). This approach is supported by Treffers-Daller et al. (2018).

Lexical sophistication refers to the proportion of relatively rare vocabulary but can also be associated with the abstractness/concreteness of words (Brysbaert et al., 2014). As in Laufer and Nation (1995) and Alp et al. (2013), separate tiers of the most frequent vocabulary were used to find the percentage of advanced tokens in text. The number of tokens not representing the 1,000, 2,000, 3,000, 4,000, and 5,000 top-frequency words according to the lemmatized version of the Estonian frequency dictionary[3] were defined. Their percentage was calculated based on the total number of tokens, not lexical tokens, mainly because the list of Estonian function words (Uiboaed, 2018) used to measure lexical density contains numerous items (1,181 out of 1,602) that do not belong to the 5,000 most frequent words and would remain undetected as rare.

Average noun abstractness was determined using an Estonian speed-reading software[4] that rates the abstractness of noun tokens based on the data from Mikk et al. (2003). The ratings are given on a three-point scale proposed by Mikk and Elts (1993), where more concrete nouns designating things or phenomena and processes perceivable by senses are respectively graded 1 and 2, while abstract nouns designating objects and notions imperceptible by senses are graded 3.

**Surface features** ($n = 9$) not requiring deeper linguistic analysis included word, sentence, and syllable count, average word length in characters and sentence length in words, the proportion of polysyllabic words (i.e., words with three or more syllables), and the LIX, SMOG, and Flesch-Kincaid grade level text complexity index. The three difficulty measures were chosen as they have proven to distinguish Estonian L2 writings with varying proficiency levels and are implemented in the ELLE writing evaluation tool.

Although it is not advised to calculate SMOG for texts shorter than 30 sentences (McLaughlin, 1969) and lexical diversity indices for texts having fewer than 100 words (e.g., Koizumi & In'nami, 2012), these features were considered potentially useful for proficiency level prediction in case of standard writing tasks with limited text length variation. The recommendation to analyze lexical diversity in random samples of equal size (e.g., Treffers-Daller et al., 2018) was disregarded, since the shortest A2-level texts only contain about 30 words.

**Error features** ($n = 9$) were based on edits proposed by either a speller or grammar corrector, which handles various types of errors such as spelling, compounding, interpunctuation, word order, word form, or lexical choice errors. The following features were extracted from the sentence-by-sentence correction input and output: percentage of words and sentences corrected by spell-checker, number of spelling corrections per sentence, average percent of speller-corrected words in a sentence, number of grammar corrections per word and sentence, percentage of grammar-corrected sentences and words occurring within corrected text spans as well as average percent of such words in a sentence. E.g., if the order of three words was changed, they were located within a word order correction. Types of corrections were only distinguished by the tool. The speller, enhanced with a list of corrections for 2,500 common spelling errors, has achieved 74% precision and 60% recall on L2 test data[5]. The grammar corrector has scored 71% in precision and 55% in recall (Luhtaru et al., 2024).

---

[3] https://www.cl.ut.ee/ressursid/sagedused1/index.php?lang=en
[4] https://kiirlugemine.keeleressursid.ee

The Pandas data analysis package for Python (McKinney, 2010) was applied to calculate features, and SPSS Statistics software for significance testing and correlation analysis. The linguistic features relevant for predicting proficiency level were chosen according to the following criteria: 1) mean values significantly distinguish at least some adjacent levels; 2) level-to-level changes are monotonic, i.e., unidirectional; 3) feature values correlate to proficiency level; 4) within-level variation between text types is non-significant or does not interfere with differentiating the levels. As an exception, non-monotonic change was allowed if no genre variation was found and the difference from all other levels was significant. A feature consistently distinguishing levels A2–C1 was also considered relevant if within-level variation affected the distinction of one level-pair but not the other two.

Welch's ANOVA was used to detect significant between-level differences, applying Bonferroni correction to control the false positive value in the multiple-comparison setting. The initial value of $\alpha = 0.05$ was divided by the total number of features (148), and the null hypothesis was rejected if $p \leq 0.0003$. The same significance level was used to determine per-level text type differences, employing the *T*-test (not assuming equal variances) or Welch's ANOVA. The Games-Howell post-hoc test with $p \leq 0.05$ was applied to define which levels (or B2-level text types) are specifically distinguished by a feature. However, despite a significant difference between adjacent levels, there may be no correlation to proficiency level due to high variation in feature values. The correlation was considered significant if the absolute value of Spearman's $\rho$ was at least 0.2, indicating a weak relationship.

### 3.3. Classification and evaluation

Machine learning models for text-level prediction were built and evaluated with the Scikit-learn Python library (Pedregosa et al., 2011)[6]. The classification pipelines comprised standardizing data with the *StandardScaler* function, using a feature selection algorithm, training a classifier, and validating it on test material. Separate pipelines were created for lexical, morphological, surface, and error features. In each case, two conditions were compared, allowing the predictive features to be chosen from 1) those regarded as relevant proficiency level predictors based on linguistic analysis, or 2) all available features.

Initial experiments with various multiclass classification algorithms highlighted seven methods to be used in the pipelines. These were logistic regression (LR), logistic regression with built-in cross-validation (LR-CV), support vector machine (SVM) using the *SVC* method, random forest (RF), multi-layer perceptron (MLP), and linear and quadratic discriminant analysis (LDA and QDA). The default hyperparameters were chosen, only increasing the number of data processing iterations when needed. With each classifier, both univariate (*SelectKBest*) and sequential forward selection (*SequentialFeatureSelector*) were applied. The first involved calculating independent feature scores (ANOVA *F*-values) and using *k* highest-scoring features to train classifiers. The second method estimated the best feature to add to the model based on five-fold cross-validation. Sequential selection was stopped if three consecutive features did not improve performance.

---

[5] The test results can be viewed at https://github.com/tlu-dt-nlp/Spell-testing.
[6] The source code for feature extraction and text classification is available at https://github.com/tlu-dt-nlp/Estonian-CEFR-Assessment.

10-fold cross-validation was used on the training set to identify the best classification parameters based on accuracy as well as the macro-average precision, recall, and F1-score per proficiency level. For each combination of feature selection and classification methods, the smallest number of features that entailed the highest possible accuracy was preferred. The five best-performing pipelines were typically chosen for evaluation on test set 1. In addition to other metrics, the recall per text type at levels A2–B2 was considered to avoid inconsistent classification. The two or three highest-scoring models were further tested for generalizability on test set 2. Balanced accuracy, i.e., macro-average recall per level, was calculated to compensate for the imbalanced dataset. In the end, different groups of relevant features were combined to train unified prediction models.

The models that best generalized to test set 2 were chosen for feature importance analysis. The contribution of individual features to prediction accuracy was measured by calculating permutation feature importance on both test sets. This model-agnostic approach is based on randomly shuffling the values of a certain feature and observing the change in model performance. Permutation was repeated 10 times for each feature to define the average decrease in classification accuracy (balanced accuracy in case of test set 2).

## 4. Results

### 4.1 Lexical features

#### 4.1.1 Relevant predictors of proficiency level

Out of 20 lexical features, seven could be considered relevant for distinguishing A2–C1-level writings of Estonian proficiency examinations. Five of them are lexical diversity measures (see Table 2). The number of lemmas, as well as the RTTR, CVV, and MTLD indices, capture a steady increase in lexical richness, although MTLD may not differentiate level A2 from B1 due to variation between text types. The mean value of MTLD in A2 personal letters (83.1, *SD* = 47.45) approaches that of level B1. Lemma count and RTTR are the most strongly correlated with the proficiency level (see Table 1 in Appendix 1).

The only PoS-specific diversity measure in the selection, adverb TTR, distinguishes levels A2–B1. The drop in the proportion of unique adverbs relates to the increase in adverb percentage at B1 (see 4.2.1). As adverb percentage and TTR do not change at levels B2–C1, the number of adverb lemmas and tokens grows at a similar rate.

The two remaining features are lexical sophistication measures. Average noun abstractness level rises consistently, whereas the use of rare vocabulary, as compared to the 5,000 most frequent words (henceforth abbreviated as rare-5,000), expands at level C1.

**Table 2.** Mean values and level-to-level differences of relevant lexical features

| Feature | Mean value per proficiency level (SD) | | | | Significant differences | | |
|---|---|---|---|---|---|---|---|
| | A2 | B1 | B2 | C1 | A2-B1 | B1-B2 | B2-C1 |
| lemma count | 33.5 (7.8) | 64.9 (11.0) | 99.6 (18.0) | 147.3 (23.4) | ✓ | ✓ | ✓ |
| root type-token ratio (RTTR) | 4.7 (0.6) | 5.9 (0.7) | 7.2 (0.8) | 9.1 (1.0) | ✓ | ✓ | ✓ |
| corrected verb variation (CVV) | 1.6 (0.4) | 2.1 (0.4) | 2.45 (0.4) | 2.9 (0.4) | ✓ | ✓ | ✓ |
| adverb type-token ratio (TTR) | 0.880 (0.162) | 0.731 (0.145) | 0.758 (0.117) | 0.732 (0.101) | ✓ | | |
| MTLD index | 74.3 (44.8) | 89.3 (35.3) | 138.6 (49.0) | 239.3 (74.4) | (✓) | ✓ | ✓ |
| avg. noun abstractness (1–3) | 1.5 (0.2) | 1.7 (0.3) | 1.9 (0.3) | 2.2 (0.2) | ✓ | ✓ | ✓ |
| words not among the most frequent 5,000 (%) | 4.0 (3.3) | 3.2 (2.1) | 3.6 (2.1) | 6.2 (3.2) | | | ✓ |

Calculating vocabulary range in comparison with the 1,000–4,000 most frequent words (rare-1,000–rare-4,000) entails non-monotonic changes between proficiency levels: feature values decrease at level B1 and increase again at C1. This can be connected to similar changes in noun percentage and lexical density, affected by the writing task and an increasing use of conjunctions as well as adverbial function words at levels B1–B2 (see section 4.2.1 and Allkivi-Metsoja, 2021). TTR, Maas, and Uber indices also change non-monotonically, suggesting that the latter two may be more sensitive to the shortness of A2-level texts than RTTR, CVV, and MTLD.

### 4.1.2 Text classification

Various classification pipelines were compared on two feature sets: lexical features selected as relevant predictors of proficiency level (LexRel) and all lexical features (LexAll). A maximum of 10 features (half of all the features) could be chosen for the models from LexAll. Based on 10-fold cross-validation, the five best-performing pipelines were determined for each feature set. The best classification parameters yielded a mean accuracy of 0.907–0.917 with LexRel and 0.908–0.920 with LexAll features. The results obtained on test set 1 were rather similar. Tables 3 and 4 highlight the best accuracy and average precision, recall, and F1 as well as the smallest standard deviations.

The LexRel models correctly classified 104–107 texts out of 120. The highest accuracy was accomplished by the RF classifier using all seven relevant features (henceforth LexRel-RF-7). MLP with three features (lemma count, noun abstractness, and rare-5,000) chosen through sequential feature selection (LexRel-MLP-sfs-3) produced smaller variation in precision but larger variation in recall. LR-CV using the five best features based on univariate selection (LexRel-LR-CV-kbest-5) attained the smallest recall and F1 variation. The features included lemma count, noun abstractness, RTTR, MTLD, and CVV. These three models accurately classified 27 A2-, 22–24 B1-, 27 B2-, and 27–30 C1-level texts. The classification error remained within one level, e.g., A2 or B2 in the case of B1. At levels A2–B2, the number of correct classifications per text type differed by 0–2, not revealing a significant bias.

**Table 3.** Test set 1 performance of classification models with relevant lexical features

| Classifier | Feature selection | No. of features | Accuracy | Precision (SD) | Recall (SD) | F1-score (SD) |
|---|---|---|---|---|---|---|
| RF | - | 7 | **0.892** | **0.893** (0.059) | **0.893** (0.061) | **0.893** (0.057) |
| MLP | sequential | 3 | 0.883 | 0.883 (**0.040**) | 0.883 (0.097) | 0.883 (0.065) |
| LR-CV | univariate | 5 | 0.875 | 0.878 (0.058) | 0.875 (**0.043**) | 0.876 (**0.039**) |
| LR-CV | univariate | 6 | 0.875 | 0.875 (0.060) | 0.878 (0.072) | 0.876 (0.062) |
| LR-CV | sequential | 3 | 0.867 | 0.865 (0.057) | 0.868 (0.087) | 0.866 (0.069) |

**Table 4.** Test set 1 performance of classification models allowing for all lexical features

| Classifier | Feature selection | No. of features | Accuracy | Precision (SD) | Recall (SD) | F1-score (SD) |
|---|---|---|---|---|---|---|
| SVM | sequential | 4 | **0.917** | 0.918 (**0.035**) | **0.918** (0.049) | **0.918** (0.037) |
| SVM | sequential | 5 | **0.917** | 0.918 (0.040) | 0.915 (0.061) | 0.916 (0.045) |
| LR | univariate | 10 | **0.917** | **0.920** (0.064) | 0.915 (**0.026**) | **0.918** (0.034) |
| SVM | sequential | 6 | 0.908 | 0.910 (**0.035**) | 0.908 (0.064) | 0.909 (0.036) |
| SVM | univariate | 7 | 0.90 | 0.910 (0.080) | 0.898 (0.041) | 0.904 (0.041) |

Model accuracy was marginally higher using LexAll features, with 108–110 texts classified correctly. The number of accurate predictions by the top three models varied between 27–28 at level A2, 25–26 at B1, 27–28 at B2, and 28–30 at C1. Within levels, correct classifications per text type differed by 0–4. The LR model with the 10 highest-ranked features (LexAll-LR-kbest-10) had the most uneven recall for B1 narrative texts and personal letters (15 *vs.* 11 classified as B1).

SVM classifiers with four and five sequentially selected features (LexAll-SVM-sfs-4 and LexAll-SVM-sfs-5) made use of Uber index and rare-3,000 in addition to lemma count, noun abstractness, and RTTR (in the five-feature model). The LexAll-LR-kbest-10 model also involved CVV and MTLD as relevant features, and the following unreliable features: rare-1,000, TTR, and conjunction, noun, and verb TTR. Some of the problematic features change non-monotonically and exhibit significant within-level differences: Uber index and rare-3,000 vary at level A2, being higher in personal letters; TTR, rare-1,000, and rare-3,000 vary at level B2, having the highest value in semi-formal letters. Conjunction, noun, and verb TTR decrease monotonically but vary by text type and do not reliably distinguish adjacent levels.

The three best LexRel and LexAll models were validated on test set 2 (see Tables 5–6). Models containing more reliable features showed better generalizability in classifying B1- and B2-level texts. With other models, the recall of B2 text types differed significantly (LexAll-LR-kbest-10) or was low for level B1 (LexAll-SVM-sfs-5 and LexAll-SVM-sfs-4).

**Table 5.** Test set 2 performance of classification models with relevant lexical features

| Model | Balanced accuracy (SD) | Recall per level and text type | |
|---|---|---|---|
| LexRel-LR-CV-kbest-5 | **0.766** (**0.173**) | A2 narrative texts | **0.983** |
| | | B1 narrative texts | 0.771 |
| | | B2 argumentative texts | **0.825** |
| | | B2 semi-formal letters | 0.794 |
| | | C1 argumentative texts | **0.50** |
| LexRel-MLP-sfs-3 | 0.756 (0.185) | A2 narrative texts | 0.965 |
| | | B1 narrative texts | **0.798** |
| | | B2 argumentative texts | 0.794 |
| | | B2 semi-formal letters | **0.810** |
| | | C1 argumentative texts | 0.458 |
| LexRel-RF-7 | 0.729 (0.191) | A2 narrative texts | 0.974 |
| | | B1 narrative texts | 0.743 |
| | | B2 argumentative texts | 0.730 |
| | | B2 semi-formal letters | 0.794 |
| | | C1 argumentative texts | 0.438 |

**Table 6.** Test set 2 performance of classification models allowing for all lexical features

| Model | Balanced accuracy (SD) | Recall per level and text type | |
|---|---|---|---|
| LexAll-LR-kbest-10 | **0.751** (0.180) | A2 narrative texts | **0.983** |
| | | B1 narrative texts | **0.789** |
| | | B2 argumentative texts | **0.873** |
| | | B2 semi-formal letters | 0.635 |
| | | C1 argumentative texts | 0.479 |
| LexAll-SVM-sfs-4 | 0.720 (**0.171**) | A2 narrative texts | 0.974 |
| | | B1 narrative texts | 0.587 |
| | | B2 argumentative texts | 0.778 |
| | | B2 semi-formal letters | **0.778** |
| | | C1 argumentative texts | **0.542** |
| LexAll-SVM-sfs-5 | 0.72 (**0.171**) | A2 narrative texts | 0.974 |
| | | B1 narrative texts | 0.615 |
| | | B2 argumentative texts | 0.778 |
| | | B2 semi-formal letters | 0.762 |
| | | C1 argumentative texts | 0.521 |

The LexRel models' classification results were closest to test set 1 for levels A2 and B1. Recall was lower for level B2 and poor for C1. It is not surprising, given that the C1-level texts of test set 2 rather resemble B2-level texts of the training set based on the RTTR ($M = 7.9$, $SD = 0.9$), CVV ($M = 2.5$, $SD = 0.4$), MTLD ($M = 156.35$, $SD = 51.5$), noun abstractness ($M = 1.95$, $SD = 0.2$), and rare-5,000 value ($M = 4.2$, $SD = 1.9$). The mean length of C1 texts is very similar in the two text samples (see 4.3.2), but the somewhat smaller lemma counts in test set 2 ($M = 130.1$, $SD = 24.2$) indicate lower lexical diversity, which may also affect lexical sophistication.

The LexRel-LR-CV-kbest-5 model that showed the best generalization abilities was analyzed for feature importance. The predictions were predominantly affected by lemma count, with the permutation score of 0.60 on test set 1 and 0.51 on test set 2 (see Figures 1–2 in Appendix 2). Noun abstractness, MTLD, and RTTR also contributed to classification accuracy, although RTTR was less significant in the case of test set 2. Omitting CVV, which received a slightly negative score on test set 2, could improve the performance. It would have to be tested, since a strong correlation between the lexical diversity features (Pearson $r > 0.7$) may reduce their permutation scores (Permutation feature importance, 2025).

## 4.2 Morphological features

### 4.2.1 Relevant predictors of proficiency level

32 relevant features were selected from the 89 morphological features that distinguish adjacent proficiency levels. These include four PoS, nominal, and verb features, five noun features, eight adjective features, and seven pronoun features (see Table 7).

**Table 7.** Mean values and level-to-level differences of relevant morphological features (non-monotonic changes are marked by an asterisk)

| Feature | Mean value per proficiency level (SD) | | | | Significant differences | | |
|---|---|---|---|---|---|---|---|
| | A2 | B1 | B2 | C1 | A2-B1 | B1-B2 | B2-C1 |
| PoS features | | | | | | | |
| adverbs (%) | 8.4 (5.1) | 11.7 (3.3) | 12.0 (3.2) | 11.4 (2.7) | ✓ | | |
| conjunctions (%) | 5.5 (3.2) | 8.1 (2.8) | 9.8 (2.2) | 8.95* (1.7) | ✓ | ✓ | ✓ |
| proper nouns (%) | 4.8 (4.5) | 3.3 (3.4) | 1.6 (1.8) | 1.1 (1.2) | ✓ | ✓ | |
| postpositions among adpositions (%) | 56.9 (45.4) | 56.5 (42.6) | 83.1 (28.6) | 77.8 (25.9) | | ✓ | |
| Nominal features | | | | | | | |
| number of cases | 6.3 (1.4) | 8.0 (1.2) | 9.2 (1.5) | 10.5 (1.2) | ✓ | ✓ | ✓ |
| nominative forms (%) | 50.3 (11.5) | 50.9 (9.3) | 42.8 (7.8) | 34.3 (7.3) | | ✓ | ✓ |
| translative forms (%) | 0.1 (0.6) | 0.3 (0.9) | 1.5 (1.8) | 3.5 (2.4) | | ✓ | ✓ |
| plural forms (%) | 8.1 (7.9) | 18.0 (9.5) | 23.3 (12.0) | 32.3 (7.5) | ✓ | ✓ | ✓ |
| Noun features | | | | | | | |
| number of cases | 5.25 (1.3) | 7.0 (1.3) | 8.15 (1.5) | 9.9 (1.3) | ✓ | ✓ | ✓ |
| nominative forms (%) | 42.3 (16.9) | 39.4 (12.9) | 35.7 (10.0) | 29.35 (6.9) | | ✓ | ✓ |
| allative forms (%) | 0.8 (2.6) | 1.8 (2.7) | 2.6 (3.1) | 2.9 (2.2) | ✓ | | |
| translative forms (%) | 0.02 (0.3) | 0.4 (1.3) | 1.6 (2.3) | 3.1 (2.4) | ✓ | ✓ | ✓ |
| plural forms (%) | 5.8 (7.7) | 15.8 (11.2) | 22.6 (13.3) | 32.2 (8.2) | ✓ | ✓ | ✓ |
| Adjective features | | | | | | | |
| number of cases | 1.4 (0.9) | 2.1 (0.9) | 3.0 (1.2) | 5.3 (1.4) | ✓ | ✓ | ✓ |
| genitive forms (%) | 2.6 (11.2) | 4.5 (8.9) | 9.3 (10.7) | 13.7 (8.6) | | ✓ | ✓ |
| partitive forms (%) | 9.8 (20.0) | 8.9 (13.7) | 7.3 (8.8) | 13.5 (9.3) | | | ✓ |
| inessive forms (%) | 0.4 (3.1) | 1.4 (5.6) | 1.5 (4.1) | 4.7 (6.4) | | | ✓ |
| elative forms (%) | 0.2 (2.0) | 0.7 (3.1) | 0.75 (5.7) | 2.5 (4.2) | | | ✓ |
| translative forms (%) | 0.4 (3.2) | 0.7 (3.2) | 2.7 (6.3) | 5.8 (5.8) | | ✓ | ✓ |
| singular forms (%) | 83.6 (32.2) | 80.5 (21.5) | 72.3 (21.4) | 61.7 (11.75) | | ✓ | ✓ |
| plural forms (%) | 5.0 (15.2) | 16.4 (19.1) | 19.6 (20.35) | 29.0 (11.7) | ✓ | | ✓ |
| Pronoun features | | | | | | | |
| number of cases | 3.2 (1.2) | 4.7 (1.1) | 6.1 (1.3) | 7.1 (1.3) | ✓ | ✓ | ✓ |
| inessive forms (%) | 0.3 (3.0) | 0.4 (1.4) | 1.7 (2.7) | 2.5 (3.3) | | ✓ | |
| elative forms (%) | 0.04 (0.5) | 0.6 (1.7) | 1.7 (2.5) | 4.5 (4.0) | ✓ | ✓ | ✓ |
| comitative forms (%) | 0.8 (2.8) | 1.8 (3.2) | 1.9 (2.7) | 2.9 (3.0) | ✓ | | ✓ |
| personal pronouns (%) | 80.5 (21.6) | 71.5 (13.9) | 48.9 (17.4) | 21.8 (12.2) | ✓ | ✓ | ✓ |
| demonstrative pronouns (%) | 10.6 (14.8) | 13.2 (9.7) | 23.95 (10.9) | 37.2 (10.4) | | ✓ | ✓ |
| interrogative-relative pronouns (%) | 0.5 (2.6) | 3.0 (5.2) | 6.3 (5.3) | 12.9 (7.8) | ✓ | ✓ | ✓ |
| Verb features | | | | | | | |
| finite forms (%) | 81.8 (13.4) | 80.9 (10.0) | 67.8 (7.4) | 65.6 (7.2) | | ✓ | |
| singular forms (%) | 60.8 (18.25) | 57.7 (14.5) | 43.4 (11.2) | 36.8 (8.6) | | ✓ | ✓ |
| components of negative forms (%) | 2.2 (4.6) | 3.15 (3.69) | 8.2 (4.4) | 4.7* (3.3) | | ✓ | ✓ |
| gerund forms (%) | 0.3 (1.5) | 0.4 (2.2) | 0.8 (1.35) | 1.6 (2.0) | | | ✓ |

The use of conjunctions, which is related to the complexity of sentence structure, was considered a reliable predictive feature despite non-monotonic change. The decreasing percentage at level C1 may be associated with growing sentence length (see 4.3.1), while the difference from B1 remains significant, and no within-level variation occurs between text types. Adverbs are more widely used from level B1, and postpositions begin to dominate over prepositions at level B2. The drop in proper noun usage reflects the shift from more personal to abstract topics, also relating to increased text length and lexical diversity.

The total number of cases, as well as the diversity of case forms in nouns, adjectives, and pronouns, increases consistently. All adjacent levels are further distinguished by the growing proportion of plural forms in nominal words, and specifically nouns. A consistent decrease in personal pronouns contrasts with increasing use of demonstrative and interrogative-relative pronouns, characteristic of more advanced noun phrases and sentence construction. These are among the morphological features that correlate most strongly with text proficiency level (see Table 2 in Appendix 1).

The most basic nominative case forms become less frequent both in total and in nouns, while other cases become more commonly used. The percentage of translative forms increases in total, in nouns, and in adjectives. Use of adjectives and pronouns in the inessive and elative cases, together with allative noun forms, genitive and partitive adjective forms, and comitative pronoun forms, also indicates significant between-level differences.

Relevant verb features comprise the dropping proportion of finite and singular verb forms, gerund forms becoming more frequent at level C1, and components of negative verb forms[7], changing non-monotonically without variation between text types. The use of verb negation is most extensive at B2, decreases at C1, and nonetheless distinguishes C1 texts from A2 and B1 texts.

The excluded distinguishing features mostly exhibit monotonic changes based on overall mean scores but fail to differentiate between levels when text type variation is considered. Some monotonic features, like essive and abessive case forms, were discarded due to very low frequency, not exceeding 0.4% even at level C1.

---

[7] The verb is typically preceded by the negative particle *ei* which is also tagged as a verb.

**4.2.2 Text classification**

Five best-performing classification pipelines that used the set of morphological features chosen as relevant proficiency level predictors (MorphRel) obtained a cross-validation accuracy of 0.855–0.868. When using all morphological features delimiting some adjacent levels (MorphAll), maximally 45 features (approx. half of the total 89) could be selected for the models. The best pipelines reached a cross-validation accuracy of 0.895–0.922. Validation on test set 1 produced slightly higher accuracy for some models. As seen in Tables 8–9, univariate feature selection and a higher number of features generally helped to achieve somewhat better classification results.

**Table 8.** Test set 1 performance of classification models with relevant morphological features

| Classifier | Feature selection | No. of features | Accuracy | Precision (SD) | Recall (SD) | F1-score (SD) |
|---|---|---|---|---|---|---|
| SVM | univariate | 26 | **0.9** | **0.905 (0.042)** | **0.90 (0.083)** | **0.903 (0.028)** |
| RF | univariate | 23 | 0.875 | 0.888 (0.076) | 0.875 (0.124) | 0.881 (0.042) |
| LDA | univariate | 25 | 0.867 | 0.878 (0.074) | 0.868 (0.087) | 0.873 (0.050) |
| RF | sequential | 17 | 0.842 | 0.850 (0.067) | 0.840 (0.085) | 0.845 (0.051) |
| QDA | sequential | 14 | 0.825 | 0.830 (0.070) | 0.825 (0.091) | 0.828 (0.058) |

**Table 9.** Test set 1 performance of classification models allowing for all morphological features

| Classifier | Feature selection | No. of features | Accuracy | Precision (SD) | Recall (SD) | F1-score (SD) |
|---|---|---|---|---|---|---|
| RF | univariate | 41 | **0.95** | **0.958** (0.058) | **0.95** (0.05) | **0.954** (0.020) |
| RF | univariate | 36 | 0.942 | 0.948 (0.044) | 0.943 (0.049) | 0.945 (**0.017**) |
| SVM | univariate | 38 | 0.933 | 0.938 (**0.035**) | 0.933 (0.046) | 0.935 (0.026) |
| SVM | univariate | 40 | 0.925 | 0.928 (0.045) | 0.925 (0.048) | 0.926 (0.029) |
| RF | sequential | 16 | 0.925 | 0.925 (0.036) | 0.925 (**0.029**) | 0.925 (0.031) |

The tested MorphRel models correctly classified 99–108 texts. SVM with 26 features (MorphRel-SVM-kbest-26) had the best accuracy as well as the most even precision, recall, and F1-score across levels. The second model chosen for further validation on test set 2 used LDA with 25 features (MorphRel-LDA-kbest-25). Although the RF classifier employing 23 features had a similarly high accuracy, its per-level recall varied the most, with only 20 B2-level texts classified accurately. MorphRel-SVM-kbest-26 and MorphRel-LDA-kbest-25 determined the correct class of 26–28 A2-, 27 B1-, 22–23 B2-, and 29–30 C1-level texts. The classification error did not exceed one level. Accurate predictions per text type differed by 0–2 for level A2, and by 1 for B1 and B2.

     The two models contained the following overlapping features: PoS features – use of adverbs, conjunctions, proper nouns, and postpositions; nominal features – number of cases, nominative, translative, and plural forms; noun features – number of cases, nominative, translative,

and plural forms; adjective features – number of cases, genitive, translative, singular, and plural forms; pronoun features – number of cases, elative forms, personal, demonstrative, and interrogative-relative pronouns; verb features – finite, singular, and negative forms. MorphRel-SVM-kbest-26 also included inessive pronoun forms.

MorphAll models accurately classified 111–114 texts. The top three models were RF classifiers using 41 (MorphAll-RF-kbest-41) and 36 features (MorphAll-RF-kbest-36), along with the SVM using 38 features (MorphAll-SVM-kbest-38). Correct predictions were made for 27–28 A2, 28–30 B1, 26–27 B2, and 30 C1 writings. While recall was rather evenly distributed across levels, it could vary between B2-level text types. MorphAll-RF-kbest-41 and MorphAll-RF-kbest-36 correctly classified 7 semi-formal letters, 10/9 personal letters, and 10 argumentative texts.

Besides the relevant features, 16 overlapping unreliable features emerged in the models: use of compound words, interjections, pronouns, and nouns; genitive and elative forms of nominal words; genitive noun forms; nominative, translative, and plural pronoun forms, indefinite pronouns; indicative, 1st person, passive, and participle verb forms, and non-finite forms in total. Additionally, the percentages of numerals and subordinating conjunctions were contained in MorphAll-SVM-kbest-38 and MorphAll-RF-kbest-41. The latter further included nominative adjective forms and 2nd-person verb forms.

Some of these features are problematic due to both non-monotonic changes and within-level variation. Compound words, interjections, pronouns, subordinate conjunctions, and 2nd-person verb forms distinguish all adjacent levels, moving both up and down in frequency. The noun percentage decreases from level A2 to B1 and increases from B2 to C1. Significant text type differences mostly occur at A2 and B2, sometimes at all three levels. Other features change monotonically, but their variation between text types does not allow for a clear between-level distinction. This makes them unreliable in differentiating consecutive proficiency levels, though they may capture important developments in learner language production, such as the increasing use of genitive case forms, passive verb forms, or indefinite pronouns.

Nonetheless, MorphAll models also achieved better scores in classifying the texts of test set 2 (see Tables 10–11). MorphRel models more accurately defined A2 texts and had a similar recall at levels B1 and C1; however, they performed worse with B2-level texts, especially semi-formal letters. All in all, the models were best generalizable for defining A2 narrative and B2 argumentative texts. Compared to test set 1, their recall was considerably lower for B1 narrations, B2 semi-formal letters, and C1 argumentations.

Most features of the MorphRel models showed some differences in test set 2 compared to the training set. Non-monotonic changes in certain features (singular, genitive, and translative adjective forms, nominative noun forms, and proper nouns), which lead to confusion between non-adjacent levels, indicate the need to treat them cautiously in automated assessment. Feature importance analysis of the best generalizable MorphRel model, MorphRel-LDA-kbest-25, confirms that these features have a small or slightly negative effect on the classification of both test sets or test set 2 (see Figures 3–4 in Appendix 2). At the same time, postpositions, adverbs, interrogative-relative pronouns, elative pronoun forms, and finite and singular verb forms are used similarly in the two datasets across all proficiency levels. Elative pronoun forms and interrogative-relative pronouns are more significant in classifying test set 2, thus facilitating generalizability.

**Table 10.** Test set 2 performance of classification models with relevant morphological features

| Model | Balanced accuracy (SD) | Recall per level and text type | |
|---|---|---|---|
| MorphRel-LDA-kbest-25 | **0.672** (0.142) | A2 narrative texts | **0.913** |
| | | B1 narrative texts | **0.633** |
| | | B2 argumentative texts | 0.651 |
| | | B2 semi-formal letters | **0.508** |
| | | C1 argumentative texts | 0.563 |
| MorphRel-SVM-kbest-26 | 0.658 (**0.134**) | A2 narrative texts | 0.887 |
| | | B1 narrative texts | 0.615 |
| | | B2 argumentative texts | **0.698** |
| | | B2 semi-formal letters | 0.397 |
| | | C1 argumentative texts | **0.583** |

**Table 11.** Test set 2 performance of classification models allowing for all morphological features

| Model | Balanced accuracy (SD) | Recall per level and text type | |
|---|---|---|---|
| MorphAll-RF-kbest-41 | **0.699** (0.084) | A2 narrative texts | 0.817 |
| | | B1 narrative texts | 0.615 |
| | | B2 argumentative texts | **0.825** |
| | | B2 semi-formal letters | **0.651** |
| | | C1 argumentative texts | **0.625** |
| MorphAll-SVM-kbest-38 | 0.686 (0.100) | A2 narrative texts | **0.843** |
| | | B1 narrative texts | **0.670** |
| | | B2 argumentative texts | 0.714 |
| | | B2 semi-formal letters | 0.619 |
| | | C1 argumentative texts | 0.563 |
| MorphAll-RF-kbest-36 | 0.681 (0.098) | A2 narrative texts | 0.826 |
| | | B1 narrative texts | 0.578 |
| | | B2 argumentative texts | 0.778 |
| | | B2 semi-formal letters | **0.651** |
| | | C1 argumentative texts | 0.604 |

The most important features include plural nominal forms, personal pronouns, and the number of adjective, pronoun, and noun cases. Correlation is likely to influence the permutation scores of features such as the number of cases and the ratio of similar grammatical categories across different nominal word classes. The strongest correlations appear between plural nominals and nouns (r = 0.915), the number of nominal and noun cases (r = 0.912), and translative nominals and nouns (r = 0.889).

## 4.3 Surface features

### 4.3.1 Relevant predictors of proficiency level

All nine surface features increase monotonically; however, six were chosen as more relevant predictors for classification (see Table 12). SMOG was preferred to other complexity indices, since the LIX and Flesch-Kincaid scores may not distinguish A2- and B1-level texts due to higher values in A2 narrative texts that resemble B1-level writings. The proportion of polysyllabic words was discarded because of significant variation at levels A2 and B2, which inhibits a clear distinction between A2–B2 texts despite a consistent increase in the mean value.

**Table 12.** Mean values and level-to-level differences of relevant surface features

| Feature | Mean value per proficiency level (SD) | | | | Significant differences | | |
|---|---|---|---|---|---|---|---|
| | A2 | B1 | B2 | C1 | A2-B1 | B1-B2 | B2-C1 |
| word count | 47.0 (12.9) | 112.1 (20.7) | 183.4 (40.2) | 257.8 (43.8) | ✓ | ✓ | ✓ |
| sentence count | 7.75 (2.3) | 13.6 (3.7) | 16.2 (4.5) | 19.2 (4.1) | ✓ | ✓ | ✓ |
| syllable count | 92.9 (23.7) | 224.8 (42.0) | 390.4 (90.7) | 652.25 (107.0) | ✓ | ✓ | ✓ |
| avg. word length (characters) | 4.9 (0.4) | 5.0 (0.4) | 5.2 (0.3) | 6.3 (0.4) | | (✓) | ✓ |
| avg. sentence length (words) | 6.3 (1.6) | 8.8 (3.0) | 12.0 (3.8) | 13.75 (2.6) | ✓ | ✓ | ✓ |
| SMOG index | 9.8 (1.5) | 11.5 (1.6) | 13.6 (1.8) | 16.9 (1.5) | ✓ | ✓ | ✓ |

Syllable and word count have the strongest correlation with the proficiency level (see Table 3 in Appendix 1). Although word count varies in B2 texts according to the writing task, it does not overlap with other levels. The mean length of B2 argumentative texts is 215.8 words ($SD = 37.8$), while personal letters contain 169.5 ($SD = 27.1$) and semi-formal letters 165.0 words ($SD = 33.85$) on average. Word length is generally shorter in personal letters and does not significantly distinguish levels A2–B1. The difference in B1- and B2-level texts is also small, as the main distinction emerges between C1 and previous levels.

### 4.3.2 Text classification

The cross-validation of classification pipelines was run with two surface feature sets, but the best-performing pipelines only included features selected as relevant predictors of proficiency level. Seven parameter sets that yielded very close cross-validation accuracy (0.930–0.938) were validated on test set 1 (see Table 13).

    The models correctly classified 110–112 texts. The highest accuracy and smallest variation in precision and recall were achieved by using SVM with two features (Surf-SVM-sfs-2) and LR

with three features (Surf-LR-sfs-3) chosen sequentially. The top two models produced identical results, accurately classifying 28 A2-, 26 B1-, 28 B2-, and 30 C1- level texts. The error was limited to one level. Correct classifications per text type differed by 0–2. The Surf-SVM-sfs-2 model relied on syllable count and average word length, while Surf-LR-sfs-3 also contained word count.

Table 13. Test set 1 performance of classification models with surface features

| Classifier | Feature selection | No. of features | Accuracy | Precision (SD) | Recall (SD) | F1-score (SD) |
|---|---|---|---|---|---|---|
| SVM | sequential | 2 | **0.933** | **0.935 (0.052)** | **0.933 (0.046)** | **0.934 (0.048)** |
| LR | sequential | 3 | **0.933** | **0.935 (0.052)** | **0.933 (0.046)** | **0.934 (0.048)** |
| LR-CV/ MLP/QDA | sequential | 2 | 0.925 | 0.928 (0.059) | 0.923 (0.061) | 0.925 (0.057) |
| MLP | sequential | 3 | 0.925 | 0.928 (0.059) | 0.923 (0.061) | 0.925 (0.057) |
| MLP | univariate | 2 | 0.917 | 0.920 (0.066) | 0.915 (0.072) | 0.918 (0.064) |

Table 14. Test set 2 performance of classification models with surface features

| Model | Balanced accuracy (SD) | Recall per level and text type | |
|---|---|---|---|
| Surf-LR-sfs-3 | **0.843 (0.083)** | A2 narrative texts | **0.983** |
| | | B1 narrative texts | **0.826** |
| | | B2 argumentative texts | 0.810 |
| | | B2 semi-formal letters | **0.730** |
| | | C1 argumentative texts | **0.792** |
| Surf-SVM-sfs-2 | 0.833 (0.089) | A2 narrative texts | **0.983** |
| | | B1 narrative texts | 0.817 |
| | | B2 argumentative texts | **0.825** |
| | | B2 semi-formal letters | 0.698 |
| | | C1 argumentative texts | 0.771 |

Further validation on test set 2 (see Table 14) revealed better generalizability for classifying A2 and B1 narrative texts than B2 and C1 argumentative texts, and especially B2 semi-formal letters. The main differences between the datasets occur in average word length. In test set 2, B1 texts have a smaller word length ($M = 4.8$, $SD = 0.3$) than A2 texts ($M = 5.3$, $SD = 0.4$), which do not differ from level B2. This confirms the unclear distinction of A2–B2. Word length is also smaller at level C1 ($M = 5.9$, $SD = 0.4$) but distinguishable from previous levels. Despite the overlapping length requirements of exam writing tasks, the word and syllable counts tend to be smaller in test set 2. Only C1 texts resemble the training set in word count ($M = 261.5$, $SD = 51.6$).

Permutation analysis of the Surf-LR-sfs-3 model showed a similar importance of word and syllable count (see Figures 5–6 in Appendix 2). Word length had a smaller but still considerable effect on prediction accuracy.

## 4.4 Error features

### 4.4.1 Relevant predictors of proficiency level

Six out of nine error features were considered relevant for distinguishing proficiency levels. Two of them were obtained with the spell-checker and four with the grammar corrector (see Table 15).

**Table 15.** Mean values and level-to-level differences of relevant error features

| Feature | Mean value per proficiency level (SD) | | | | Significant differences | | |
|---|---|---|---|---|---|---|---|
| | A2 | B1 | B2 | C1 | A2-B1 | B1-B2 | B2-C1 |
| spell-corrected words (%) | 5.2 (4.7) | 4.0 (2.5) | 2.7 (1.6) | 1.8 (1.1) | | ✓ | ✓ |
| avg. % of spell-corrected words in a sentence | 4.8 (4.6) | 4.0 (2.8) | 2.6 (1.8) | 1.9 (1.3) | | ✓ | ✓ |
| grammar corrections per word | 0.038 (0.023) | 0.015 (0.010) | 0.012 (0.009) | 0.008 (0.004) | ✓ | (✓) | ✓ |
| grammar corrections per sentence | 0.249 (0.210) | 0.145 (0.254) | 0.144 (0.128) | 0.104 (0.073) | ✓ | | ✓ |
| words within grammar corrections (%) | 27.6 (10.6) | 27.6 (8.3) | 22.2 (6.2) | 15.5 (5.2) | | ✓ | ✓ |
| avg. % of words within grammar corrections in a sentence | 26.7 (10.4) | 27.2 (8.4) | 21.7 (6.5) | 16.0 (5.5) | | ✓ | ✓ |

The percentage of words corrected by the speller does not differentiate A2- and B1-level texts, as A2 personal letters ($M = 4.4\%$, $SD = 4.4$) fall closer to B1 writings. In line with this, the average proportion of spell-corrected words per sentence also changes at levels B2 and C1. On the other hand, the ratio of grammar corrections to the number of words and sentences distinguishes A2–B1 and B2–C1. Although statistically significant, the difference in the corrections-to-words ratio between B1 and B2 is marginal. The percentage of words that occur within corrected text spans and the average proportion of such words in a sentence drop from B1 to C1. The features extracted by grammatical error correction correlate more strongly with the proficiency level (see Table 4 in Appendix 1).

    The percentage of sentences with grammar corrections changes non-monotonically. The percentage of spell-corrected sentences and the ratio of spelling corrections per sentence were excluded due to text type variation at level A2 and B2, respectively.

### 3.4.2 Text classification

The best-performing classification pipelines attained with the set of all error features (ErrorAll) and relevant proficiency level predictors (ErrorRel) partially overlapped. In two of the five highest-scoring ErrorAll pipelines, only ErrorRel features got selected. These also belonged to the top five

ErrorRel pipelines. The best cross-validation accuracy (0.715) was accomplished by the MLP algorithm with two features from sequential selection (ErrorRel-MLP-sfs-2). The other overlapping pipeline, receiving the third-highest cross-validation accuracy (0.692) in both cases, used an MLP with six features from univariate selection. The three best non-overlapping ErrorAll pipelines were validated alongside five ErrorRel pipelines (see Tables 16–17).

**Table 16.** Test set 1 performance of classification models with relevant error features

| Classifier | Feature selection | No. of features | Accuracy | Precision (SD) | Recall (SD) | F1-score (SD) |
| --- | --- | --- | --- | --- | --- | --- |
| MLP | sequential | 2 | **0.725** | **0.728 (0.066)** | **0.725** (0.180) | **0.726 (0.104)** |
| RF | sequential | 2 | 0.683 | 0.693 (0.128) | 0.683 (**0.165**) | 0.688 (0.131) |
| MLP | univariate | 6 | 0.683 | 0.683 (0.127) | 0.683 (0.191) | 0.683 (0.147) |
| RF | univariate | 6 | 0.667 | 0.653 (0.096) | 0.665 (0.203) | 0.659 (0.147) |
| RF | sequential | 5 | 0.625 | 0.613 (0.091) | 0.625 (0.189) | 0.616 (0.142) |

**Table 17.** Test set 1 performance of classification models allowing for all error features

| Classifier | Feature selection | No. of features | Accuracy | Precision (SD) | Recall (SD) | F1-score (SD) |
| --- | --- | --- | --- | --- | --- | --- |
| RF | sequential | 3 | **0.7** | **0.693** (0.063) | **0.700** (0.202) | **0.696 (0.125)** |
| RF | sequential | 4 | 0.667 | 0.660 (0.092) | 0.668 (0.203) | 0.663 (0.140) |
| RF | univariate | 9 | 0.667 | 0.653 (**0.054**) | 0.668 (0.218) | 0.66 (0.135) |

The texts of test set 1 were also most accurately classified by the ErrorRel-MLP-sfs-2 model, which relied on the ratio of grammar corrections per word and sentence. The second-best ErrorRel model used RF classification with the same two sequentially selected features (ErrorRel-RF-sfs-2). While all models poorly predicted the level of B1 and B2 texts, these two provided the most even number of correct classifications per level. Altogether, 82/87 writings were categorized accurately, including 28/26 A2-, 15/18 B1-, 17/15 B2-, and 22/28 C1-level texts. When tolerating one-level error, the accuracy of both models was 0.975. Correct predictions per text type differed by 2 for A2, 3/0 for B1, and 1–4/0–3 for B2, indicating genre-based variation.

The best ErrorAll model, using RF classification with three features from sequential feature selection (ErrorAll-RF-sfs-3), correctly classified 84 texts: 14–28 per proficiency level. Recall per genre differed by 0–3 texts. In addition to the features of the best ErrorRel models, it involved the percentage of spell-corrected sentences that distinguish level C1 from B2 and B1 but not from A2 due to large variation.

The two ErrorRel models and one ErrorAll model were validated on test set 2 (see Table 18). ErrorAll-RF-sfs-3 and ErrorRel-RF-sfs-2 yielded very similar scores. They were best able to generalize for A2-level texts and B2-level argumentative writings, and produced the most different results from test set 1 in classifying C1-level texts. The ErrorRel models had a smaller recall variation between B2-level text types.

**Table 18.** Test set 2 performance of classification models with error features

| Model | Balanced accuracy (SD) | Recall per level and text type | |
|---|---|---|---|
| ErrorAll-RF-sfs-3 | **0.607** (0.197) | A2 narrative texts | **0.939** |
| | | B1 narrative texts | 0.431 |
| | | B2 argumentative texts | 0.476 |
| | | B2 semi-formal letters | **0.635** |
| | | C1 argumentative texts | **0.5** |
| ErrorRel-RF-sfs-2 | 0.6 (**0.192**) | A2 narrative texts | 0.930 |
| | | B1 narrative texts | **0.459** |
| | | B2 argumentative texts | **0.508** |
| | | B2 semi-formal letters | 0.556 |
| | | C1 argumentative texts | 0.479 |
| ErrorRel-MLP-sfs-2 | 0.572 (**0.192**) | A2 narrative texts | 0.904 |
| | | B1 narrative texts | **0.459** |
| | | B2 argumentative texts | 0.397 |
| | | B2 semi-formal letters | 0.492 |
| | | C1 argumentative texts | 0.479 |

While the ratio of grammar corrections per word is higher in B1- and C1-level texts of test set 2 compared to the training set, the grammar corrections per sentence ratio causes confusion between various levels. B1 texts and B2 argumentative texts are similar to A2 writings. In contrast, B2 semi-formal letters stand closest to C1-level texts. Such a difference between the two B2-level text types was not expected based on the training data.

Feature importance was calculated for the ErrorRel-RF-sfs-2 model, revealing that the ratio of errors per word plays a more significant role compared to the error-per-sentence ratio. The respective permutation scores were 0.44 and 0.21 on test set 1, and 0.40 and 0.16 on test set 2.

**4.5 Mixed feature set**

When all types of relevant linguistic features (51 in total) were combined in proficiency level classification, the five best-performing pipelines demonstrated a higher cross-validation accuracy (0.943–0.952) and test accuracy (see Table 19) compared to using separate feature categories.

The three best models could correctly classify 117–118 texts, 29–30 for each level. They made use of LR or SVM classification with 23–24 features obtained through univariate selection (accordingly abbreviated as Mix-LR-kbest-24, Mix-SVM-kbest-23, and Mix-SVM-kbest-24). The following features were included in the models: lexical features – lemma count, RTTR, CVV, MTLD, noun abstractness; morphological features – number of cases, nominative (only in 24-feature-models), translative and plural forms of nominal words, number of noun, adjective and pronoun cases, plural noun forms, personal, demonstrative and interrogative-relative pronouns, finite verb forms; surface features – word, sentence, and syllable count, average word and sentence length, SMOG index; error features – grammar corrections per word.

**Table 19.** Test set 1 performance of classification models with mixed relevant features

| Classifier | Feature selection | No. of features | Accuracy | Precision (SD) | Recall (SD) | F1-score (SD) |
|---|---|---|---|---|---|---|
| LR | univariate | 24 | **0.983** | **0.985 (0.015)** | **0.985** (0.015) | **0.985 (0.015)** |
| SVM | univariate | 23 | 0.975 | 0.978 (0.025) | 0.978 (**0.013**) | 0.978 (0.017) |
| SVM | univariate | 24 | 0.975 | 0.978 (0.025) | 0.978 (**0.013**) | 0.978 (0.017) |
| SVM | univariate | 27 | 0.967 | 0.970 (0.037) | 0.968 (0.025) | 0.969 (0.025) |
| RF | sequential | 12 | 0.925 | 0.928 (0.059) | 0.923 (0.061) | 0.925 (0.057) |

When validated on test set 2, the mixed feature set did not outperform surface features (see Table 20). This is likely caused by the more notable differences in morphological features between the two exam datasets, mostly affecting levels B1 and C1. However, Mix-SVM-kbest-23, which demonstrated the best generalizability, can be considered the most reliable model for overall linguistic assessment of a learner's text.

**Table 20.** Test set 2 performance of classification models with mixed relevant features

| Model | Balanced accuracy (SD) | Recall per level and text type | |
|---|---|---|---|
| Mix-SVM-kbest-23 | **0.796 (0.110)** | A2 narrative texts | 0.965 |
| | | B1 narrative texts | **0.743** |
| | | B2 argumentative texts | **0.857** |
| | | B2 semi-formal letters | 0.762 |
| | | C1 argumentative texts | **0.667** |
| Mix-SVM-kbest-24 | 0.768 (0.139) | A2 narrative texts | **0.983** |
| | | B1 narrative texts | 0.706 |
| | | B2 argumentative texts | 0.794 |
| | | B2 semi-formal letters | 0.762 |
| | | C1 argumentative texts | 0.604 |
| Mix-LR-kbest-24 | 0.750 (0.152) | A2 narrative texts | 0.965 |
| | | B1 narrative texts | 0.661 |
| | | B2 argumentative texts | 0.841 |
| | | B2 semi-formal letters | **0.778** |
| | | C1 argumentative texts | 0.563 |

As illustrated by Figures 1–2, the classification errors on both test sets involved confusion with an adjacent proficiency level. Predicting a lower level, rather than a higher level, can be seen as problematic, but is explained by the partially lower linguistic complexity of the older test data.

According to permutation feature importance, the predictions were similarly influenced by morphological and surface features (see Figures 3–4). The combined score of morphological features was 0.13 on test set 1 and 0.29 on test set 2. The surface feature scores totaled 0.16 and 0.33, respectively. Lexical features had a more modest impact, with total scores of 0.10 and 0.20.

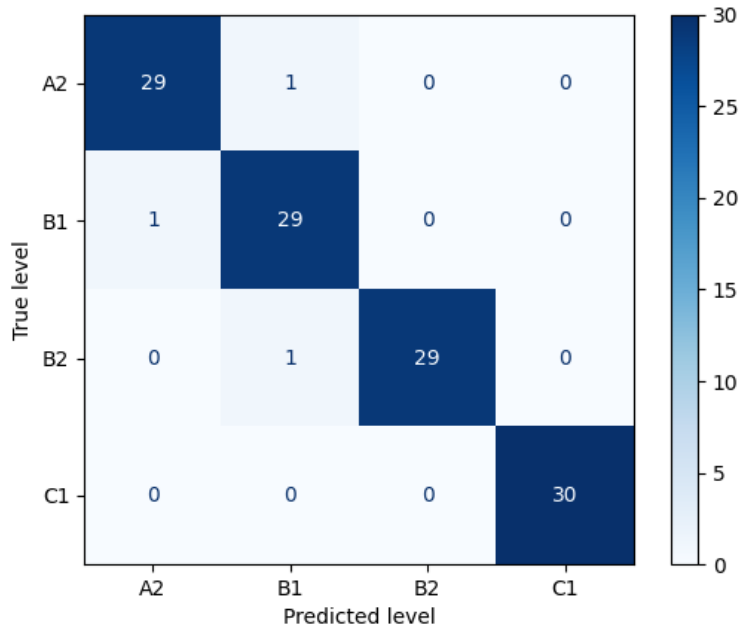

**Figure 1.** Confusion matrix of the Mix-SVM-kbest-23 model's performance on test set 1

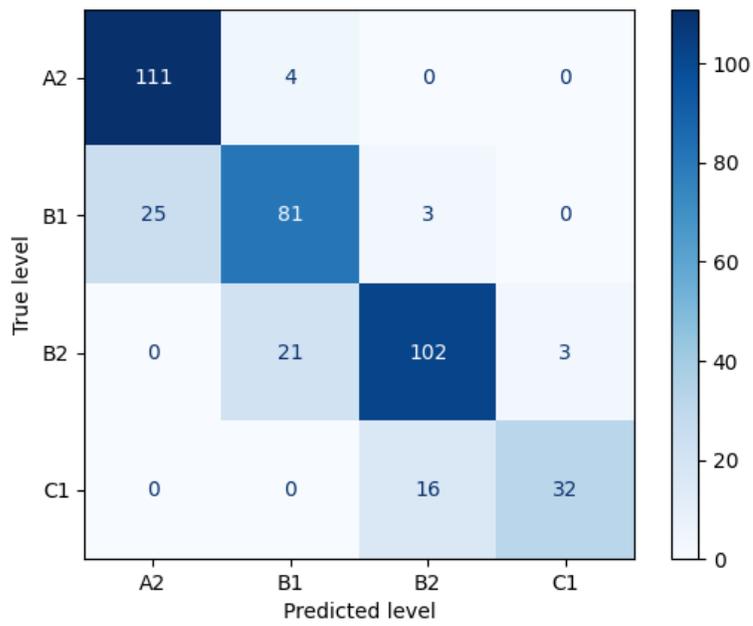

**Figure 2.** Confusion matrix of the Mix-SVM-kbest-23 model's performance on test set 2

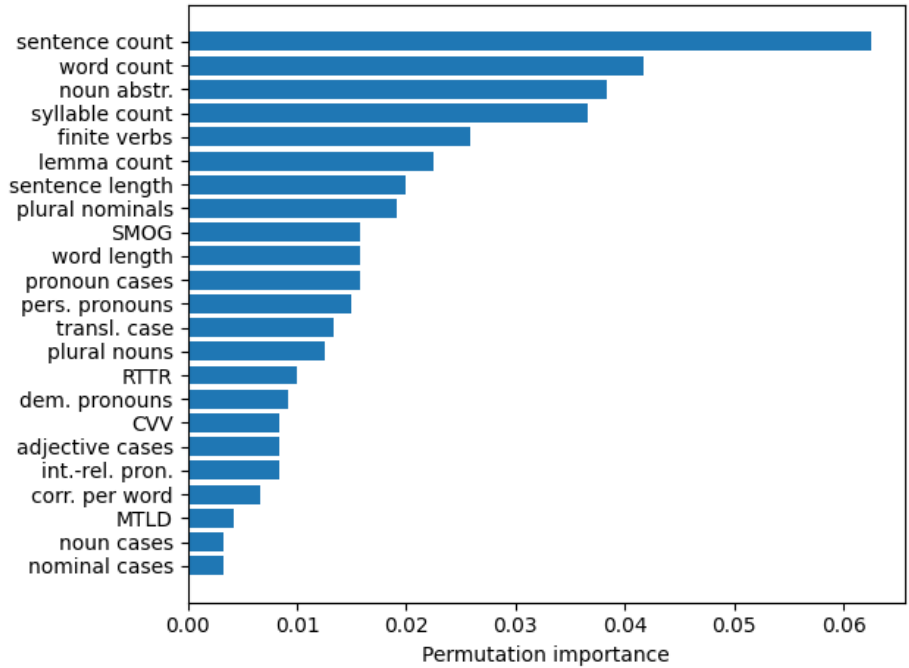

**Figure 3.** Permutation feature importances of the Mix-SVM-kbest-23 model calculated on test set 1

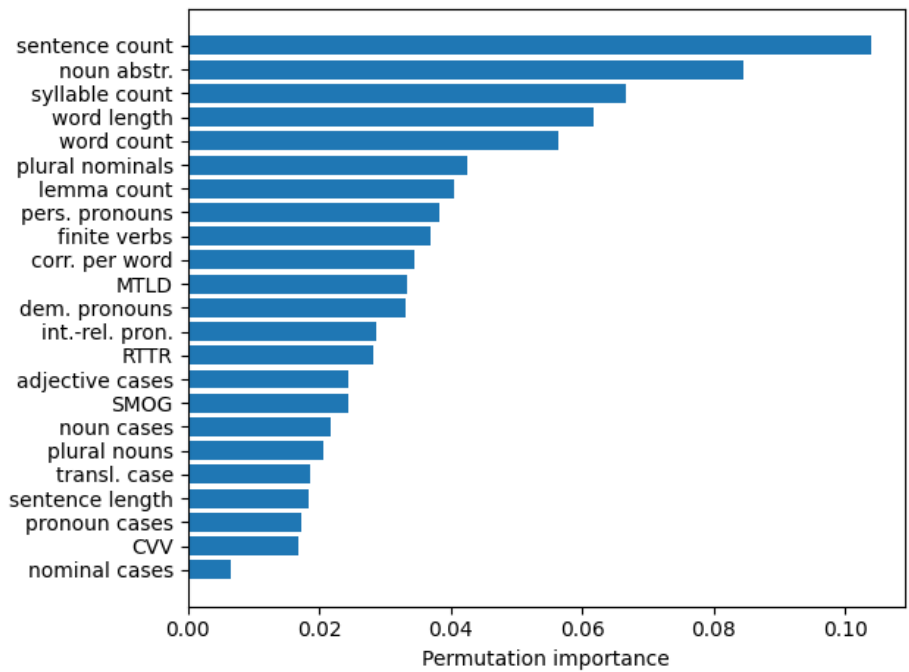

**Figure 4.** Permutation feature importances of the Mix-SVM-kbest-23 model calculated on test set 2

Sentence, word, and syllable count, noun abstractness and lemma count, as well as finite verb forms and plural nominal forms emerged as the most important individual features. Sentence length and SMOG index proved more significant for classifying test set 1. Word length, MTLD, corrections-per-word ratio, and the use of different pronoun types were more important for classifying test set 2 and may thus be useful for ensuring model generalizability. The importance of the number of pronoun, adjective, and noun case forms varied across test sets. The general number of nominal cases in the text, which is correlated with these three features and has the smallest effect on prediction accuracy, could likely be omitted without reducing performance.

In conclusion, the mixed text assessment model relies on a diverse set of features. Alongside superficial textual complexity measures and error frequency, its proficiency level predictions account for various aspects of grammatical and lexical complexity, including lexical richness and sophistication, nominal word and verb inflection.

## 5. Conclusions

This work combined CEFR-based comparative analysis and classification of Estonian learner writings by first defining the relationship between various linguistic features and text level. About one-third of the observed lexical and morphological features and most surface and error features were regarded as relevant for proficiency assessment based on the criteria of distinction between adjacent levels, monotonic changes, correlation to text level, and small within-level variation.

Among lexical features, the diversity measures and noun abstractness proved to be better predictors of writing proficiency than rare vocabulary. Considering only content words when calculating rare word percentages would avoid dependence on lexical density that exhibits a non-monotonic change, also detected by Alp et al. (2013). Furthermore, applying learner-corpus-based frequency lists instead of those based on native-speaker reference corpora could help to distinguish proficiency levels more efficiently (e.g., Naismith et al., 2018; Pilán & Volodina, 2018).

Morphological features characterizing nominal words (including nouns, adjectives, and pronouns specifically) were found more relevant for predicting the CEFR levels than PoS and verb features, most of which were significantly task-dependent. Current results largely confirm earlier findings obtained with a 480-text subset of the main exam sample used in this study (Allkivi-Metsoja, 2021; 2022). These analyses relied on manually corrected tagging output and a less diverse text selection in terms of genre, which might explain some differences. For example, genitive nouns and nominals in general were no longer deemed reliable for differentiating consecutive levels due to within-level variation.

Like PoS-specific nominal features and lexical sophistication measures, surface and error features had previously not been used for the automated assessment of Estonian L2 texts. The grammar corrector, which recognizes various types of errors, was more useful for text comparison and classification than the spell-checker. Surface features are highly predictive of proficiency level, while variables like word and syllable count cannot be generalized to less standardized writing situations. This also applies to text-length-dependent lexical features, such as lemma count, RTTR, and CVV.

Several level-distinguishing features have proven important in the automated assessment of various languages. Besides error frequency, word and sentence count and length, the number of unique words, and the SMOG, RTTR, CVV, and MTLD indices, a few more specific overlapping features can be pointed out. An increase in lexical variation of adverbs and the frequency of plural word forms, as well as a decrease in singular and nominative case forms, are also proficiency markers of L2 German (Szügyi et al., 2019). In L2 Portuguese, subordinate and relative clauses become actively used at more advanced levels (Ribeiro-Flucht et al., 2024), which is equivalent to the increased use of conjunctions and interrogative-relative pronouns noted for Estonian. The decreasing frequency of proper nouns and personal pronouns has been found relevant for predicting L2 English proficiency (Vajjala, 2018). The use of different pronoun types has helped to predict advancement in L2 Czech (Rysová et al., 2019). This indicates a certain extent of cross-linguistic parallels in the progression of language proficiency. However, such similarities depend on the language structure, and differences are likely to occur both in feature values and the exact levels they distinguish.

The classification experiments yielded better results in predicting the proficiency level of Estonian L2 texts than previously obtained by Vajjala and Lõo (2014), although it must be noted that they used a more heterogeneous dataset of specially graded non-exam writings. The best models based on lexical, morphological, and surface features achieved an accuracy of around 0.9 on the main test set 1. With error features, accuracy was about 0.7.

Evaluation of the top-performing classifiers on external test set 2 from earlier proficiency exams revealed an increase in complexity of the writing assignment responses between 2010 and 2017–2020. This was most evident at level C1. The main differences from the training set occurred in morphological features, such as the number of case forms and their proportions among nominal words. Lexical complexity and text and word length tended to be smaller in test set 2, while B1- and C1-level texts had a higher ratio of grammar corrections per word. This phenomenon, well-received by the organizers of Estonian proficiency testing, deserves further investigation. The best classification models still exceeded 0.7 or even 0.8 in balanced accuracy, which can be considered a relatively good result.

Omitting genre-dependent distinguishing features led to rather similar accuracy compared to using all available features, but allowed better generalizability across text types. The LexRel models consistently ensured an even recall of B1 genres in test set 1, as well as a high recall of B1 texts in general and an even recall of B2 genres in test set 2. The MorphRel models showed smaller variation in the recall of B2 genres in test set 1, while per-level recall distribution was more uneven, resulting in somewhat lower accuracy than scored by the MorphAll models. Nevertheless, a considerably smaller number of features (25–26 *vs.* 36–41) may still benefit the generalizability of the MorphRel models, which needs further validation. The top-ranked surface feature pipelines only involved relevant CEFR level predictors. Two of the three best error-based classifiers also relied on relevant variables, achieving a more even recall of B2 genres in test set 2 as compared to the ErrorAll model.

Combining various types of relevant linguistic features enabled a classification accuracy of around 0.98 on test set 1 and 0.8 on test set 2. All feature subsets were represented in the mixed models. The findings that a combined selection and reduced number of features improve prediction accuracy stand in line with previous research (Pilán, 2018; Vajjala, 2018; Szügyi et al., 2019). The CEFR classification results in this study can be compared with or exceed the best performance metrics reported in earlier work: holdout test set accuracy of 0.84 for Swedish (Pilán, 2018) and 0.98 for English (Schmalz & Brutti, 2022), cross-validation accuracy of 0.82 for German (Szügyi et al., 2019) and French (Gaillat et al., 2022), and weighted F1-score of 0.84 for Italian (Vajjala & Rama, 2018). It should, however, be considered that these scores stem from three-, five-, or six-level classification scenarios with varying feature sets. The results should also be validated on a larger dataset containing more recent exam writings.

## 6. Implications and limitations

Selected classification pipelines have been integrated into the ELLE Writing Evaluator[8] to assess general linguistic proficiency and assign sub-ratings based on lexical, grammatical, and surface complexity of Estonian L2 learner texts. Error-based assessment will be improved before implementation, utilizing a more advanced grammar correction and error explanation tool under development (Vainikko et al., 2025) and taking into account specific error types. When choosing optimal parameter sets, the preference was given to pipelines that best generalized to external data: LexRel-LR-CV-kbest-5, MorphRel-LDA-kbest-25, Surf-LR-sfs-3, and Mix-SVM-kbest-23. Final models were trained on the main dataset, including test set 1. Permutation feature importance analysis has suggested the need for further experiments with removing certain low-scoring features that may be redundant for classification. Another planned addition is to generate more extensive feedback based on individual features and their relation to proficiency level. Currently, the user sees a general description of the assessment methodology and linguistic variables considered.

Such text grading functionality concerns only certain aspects of linguistic competence and disregards pragmatic and sociolinguistic competence (see CEFR, 2020), not providing a comprehensive assessment of writing proficiency. It can rather be used to obtain an indicative measure of linguistic complexity in a text, as compared to other learners' writing production, representing different CEFR levels. Such feedback, in combination with automated corrections, can assist writing practice and grading, without replacing expert evaluation.

On the other hand, large-scale statistical analysis of authentic language use rated to correspond to a particular level can lead to more evidence-based practices in teaching and testing. Elaborating on the corpus-based comparison and proficiency prediction results of this study will help to specify the CEFR descriptors for Estonian. The lexical and grammatical profile of Estonian as L2 draw on coursebook, learner, and general corpus materials (Üksik et al., 2021). At the same time, language assessment needs to rely on actual performance mapped to the proficiency scale instead of echoing the expectations set for learners.

---

[8] https://elle.tlu.ee/corrector

The main limitations of this study relate to the language material. Firstly, texts not matching the criteria of level A2 were not analyzed. Writings that should be categorized as pre-A2 would currently be classified as A2. Secondly, the sample was limited to a few text types represented at the proficiency exams. The usability of the classification models in assessing other genres and texts produced in non-exam settings should be tested separately. The possible effect of gender bias should also be explored in future research due to unequal gender representation in the training data.

While multiclass classification offers a first indication of the probable proficiency level, binary classification can lead to more accurate predictions. This could be especially useful for identifying intermediate levels tending to have lower F1-scores. Further experiments should also consider features that were hereby discarded, such as syntactic features derived from deep parsing, PoS and dependency n-grams, or discourse features characterizing cohesion. Excluding highly correlated features may improve model generalizability but also affect interpretation, as pointed out by Ribeiro-Flucht et al. (2024).

To summarize, the attempt to connect linguistic analysis and automated CEFR classification of Estonian learner texts shows promising results for placement testing, individual and classroom learning. High accuracy scores also indicate potential for developing exam-grading applications (assisting, not replacing expert graders) based on a similar approach. The framework proposed for defining relevant markers of level-to-level progress and using these features to build more explainable writing assessment tools can be adopted for different languages. In prospect, such feature-based assessment could be combined with other methods for scoring various aspects of writing proficiency, including thematic development, coherence, and cohesion. It could also provide data-driven input for prompting LLMs to generate learner feedback.

**Acknowledgements**

This study was supported by the Tallinn University Research Fund and the Ustus Agur Scholarship for doctoral students awarded by the Estonian Association of Information Technology and Telecommunications and the Estonian Education and Youth Board.

# Appendix 1. Correlations between relevant predictors and the CEFR level

Tables 1–4 present Spearman's rank correlation coefficients between the linguistic features considered relevant for predicting Estonian L2 writing proficiency and the CEFR level (A2–C1) of the texts in the training set. All correlations are significant at the 0.01 level.

**Table 1.** Lexical feature correlations with proficiency level

| Feature | Spearman $\rho$ |
|---|---|
| lemma count | 0.949 |
| root type-token ratio (RTTR) | 0.914 |
| corrected verb variation (CVV) | 0.773 |
| MTLD index | 0.757 |
| avg. noun abstractness (1–3) | 0.738 |
| adverb type-token ratio (TTR) | -0.269 |
| words not among the most frequent 5,000 (%) | 0.261 |

**Table 2.** Morphological feature correlations with proficiency level

| Feature | Spearman $\rho$ |
|---|---|
| PoS features | |
| conjunctions (%) | 0.419 |
| proper nouns (%) | -0.376 |
| postpositions among adpositions (%) | 0.364 |
| adverbs (%) | 0.270 |
| Nominal features | |
| number of cases | 0.774 |
| translative forms (%) | 0.721 |
| plural forms (%) | 0.693 |
| nominative forms (%) | -0.571 |
| Noun features | |
| number of cases | 0.787 |
| plural forms (%) | 0.705 |
| translative forms (%) | 0.648 |
| allative forms (%) | 0.437 |
| nominative forms (%) | -0.351 |

| Adjective features | |
| --- | --- |
| number of cases | 0.769 |
| genitive forms (%) | 0.568 |
| plural forms (%) | 0.561 |
| translative forms (%) | 0.525 |
| singular forms (%) | -0.497 |
| inessive forms (%) | 0.418 |
| elative forms (%) | 0.381 |
| partitive forms (%) | 0.305 |
| Pronoun features | |
| personal pronouns (%) | -0.792 |
| number of cases | 0.769 |
| interrogative/relative pronouns (%) | 0.723 |
| demonstrative pronouns (%) | 0.690 |
| elative forms (%) | 0.576 |
| inessive forms (%) | 0.430 |
| comitative forms (%) | 0.336 |
| Verb features | |
| singular forms (%) | -0.585 |
| finite forms (%) | -0.559 |
| gerund forms (%) | 0.441 |
| components of negative forms (%) | 0.395 |

**Table 3.** Surface feature correlations with proficiency level

| Feature | Spearman $\rho$ |
| --- | --- |
| syllable count | 0.955 |
| word count | 0.937 |
| SMOG index | 0.849 |
| avg. sentence length (words) | 0.791 |
| sentence count | 0.754 |
| avg. word length (characters) | 0.704 |

**Table 4.** Error feature correlations with proficiency level

| Feature | Spearman $\rho$ |
|---|---|
| grammar corrections per word | -0.705 |
| words within grammar corrections (%) | -0.509 |
| avg. % of words within grammar corrections in a sentence | -0.462 |
| grammar corrections per sentence | -0.393 |
| spell-corrected words (%) | -0.378 |
| avg. % of spell-corrected words in a sentence | -0.313 |

**Appendix 2. Feature importances of best-performing models per linguistic feature group**

Figures 1–6 present permutation importance scores of the features included in the best-performing CEFR classification models assessing specific linguistic aspects of a learner text: LexRel-LR-CV-kbest-5 for grading lexical complexity (see section 4.1.2), MorphRel-LDA-kbest-25 for grading grammatical complexity (see 4.2.2), and Surf-LR-sfs-3 for grading surface complexity (see 4.3.2). Permutation importance has been calculated based on the classification accuracy of the two test sets, using balanced accuracy for test set 2.

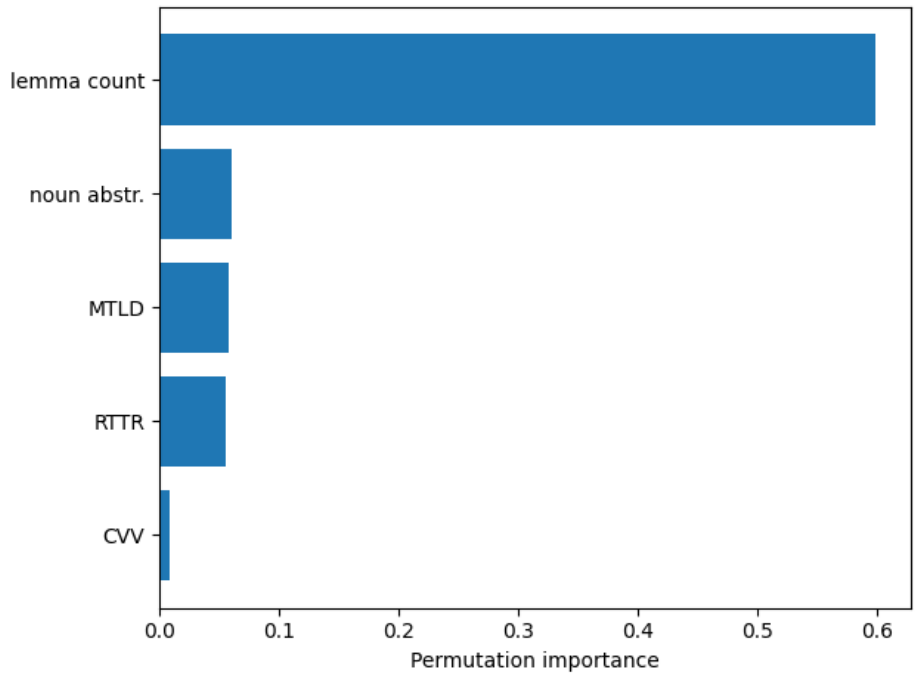

**Figure 1.** Permutation feature importances of the LexRel-LR-CV-kbest-5 model calculated on test set 1

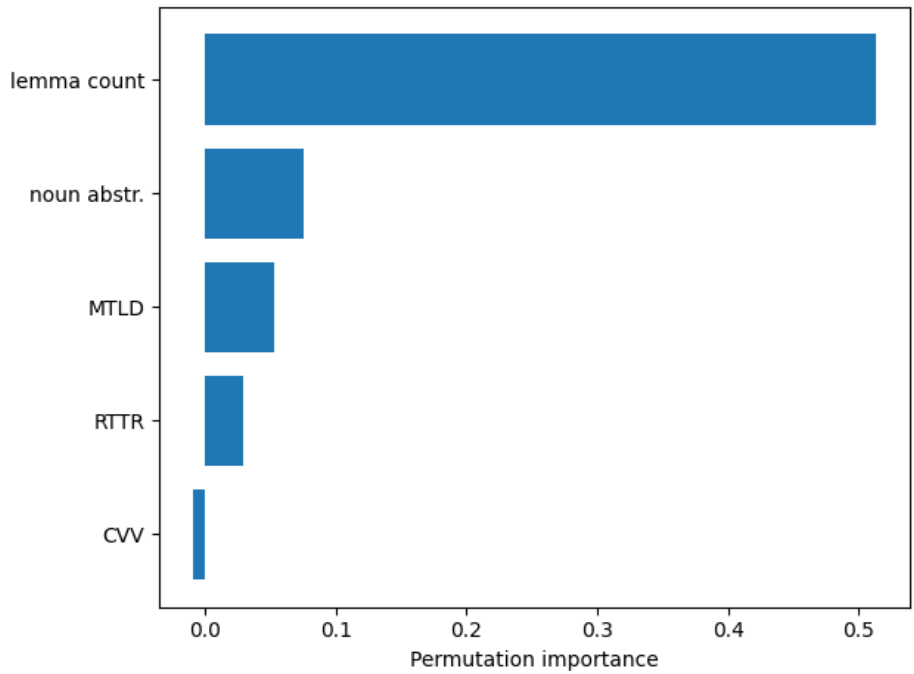

**Figure 2.** Permutation feature importances of the LexRel-LR-CV-kbest-5 model calculated on test set 2

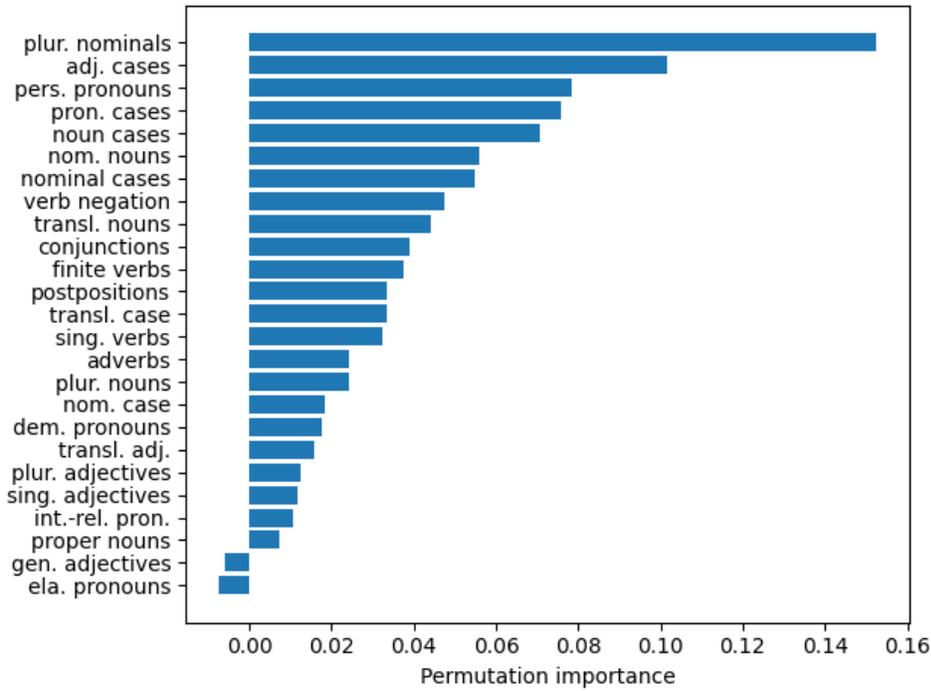

**Figure 3.** Permutation feature importances of the MorphRel-LDA-kbest-25 model calculated on test set 1

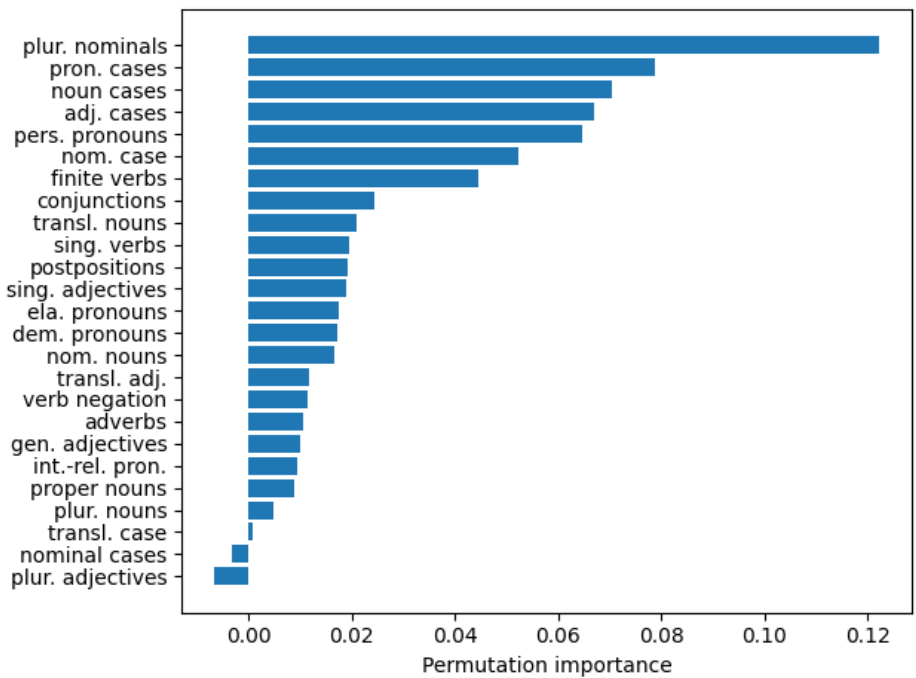

**Figure 4.** Permutation feature importances of the MorphRel-LDA-kbest-25 model calculated on test set 2

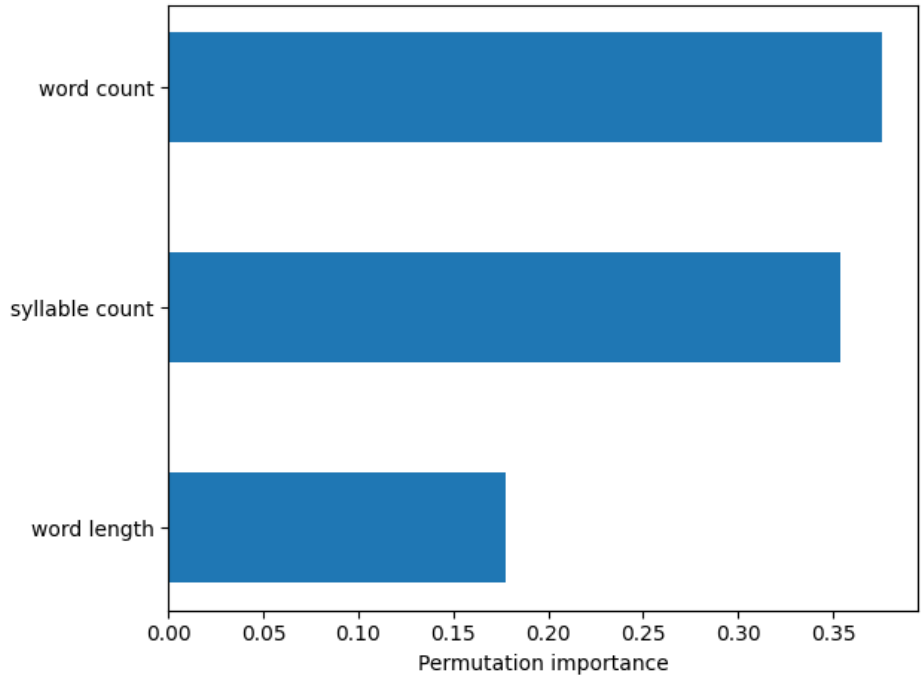

**Figure 5.** Permutation feature importances of the Surf-LR-sfs-3 model calculated on test set 1

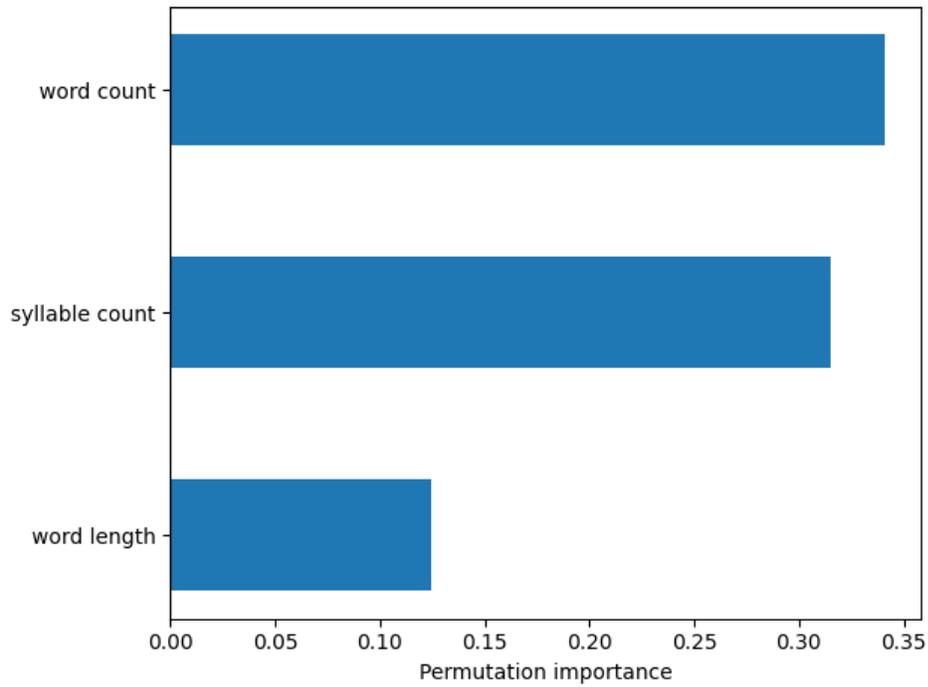

**Figure 6.** Permutation feature importances of the Surf-LR-sfs-3 model calculated on test set 2